\def\year{2021}\relax
\pdfoutput=1 %needed for arxiv
\documentclass[letterpaper]{article} % DO NOT CHANGE THIS

\usepackage{aaai21}  % DO NOT CHANGE THIS
\usepackage{times}  % DO NOT CHANGE THIS
\usepackage{helvet} % DO NOT CHANGE THIS
\usepackage{courier}  % DO NOT CHANGE THIS
\usepackage[hyphens]{url}  % DO NOT CHANGE THIS
\usepackage{graphicx} % DO NOT CHANGE THIS
\usepackage{graphicx}  % DO NOT CHANGE THIS
\usepackage{natbib}  % DO NOT CHANGE THIS AND DO NOT ADD ANY OPTIONS TO IT
\usepackage{caption} % DO NOT CHANGE THIS AND DO NOT ADD ANY OPTIONS TO IT
\usepackage[caption=false]{subfig}
\usepackage{algorithmic}
\usepackage[ruled]{algorithm2e}
\usepackage{booktabs}

\usepackage{amsmath,amsfonts,bm}

\def\eqref#1{equation~\ref{#1}}
\def\1{\bm{1}}

\DeclareMathAlphabet{\mathsfit}{\encodingdefault}{\sfdefault}{m}{sl}
\SetMathAlphabet{\mathsfit}{bold}{\encodingdefault}{\sfdefault}{bx}{n}

\newcommand{\enote}[1]{\textcolor{red}{Berry: #1}}
\newcommand{\ignore}[1]{}

\def\w{{\bf w}}

\def\b{{\bf b}}

\def\x{{\bf x}}

\title{Margin-Based Regularization and Selective Sampling in Deep Neural Networks}
\author{
    Berry Weinstein, \textsuperscript{\rm 1}
    Shai Fine, \textsuperscript{\rm 2}
    Yacov Hel-Or \textsuperscript{\rm 1}}
    
\affiliations{
    \textsuperscript{\rm 1} Efi Arazi School of Computer Science, Interdisciplinary Center, Herzliya, Israel \\
    \textsuperscript{\rm 2} The Data Science Institute, Interdisciplinary Center, Herzliya, Israel \\
    berry.weinstein@post.idc.ac.il, shai.fine@idc.ac.il, toky@idc.ac.il}

\begin{document}

\maketitle
\begin{abstract}
%Furthermore, we extend our derivation for a selective sampling scheme in DNNs in an unsupervised manner. We show that we can accelerate the training process by selecting samples according to its {\it minimal margin score} (MMS), which measures the minimal amount of displacement an input should undergo until its predicted classification is switched. 

We derive a new margin-based regularization formulation, termed {\it multi-margin regularization} (MMR), for {\it deep neural networks} (DNNs). The MMR is inspired by principles that were applied in margin analysis of shallow linear classifiers, e.g., {\it support vector machine} (SVM). Unlike SVM, MMR is continuously scaled by the radius of the bounding sphere (i.e., the maximal norm of the feature vector in the data), which is constantly changing during training. We empirically demonstrate that by a simple supplement to the loss function, our method achieves better results on various classification tasks across domains. Using the same concept, we also derive a selective sampling scheme and demonstrate accelerated training of DNNs by selecting samples according to a {\it minimal margin score} (MMS). This score measures the minimal amount of displacement an input should undergo until its predicted classification is switched. We evaluate our proposed methods on three image classification tasks and six language text classification tasks. Specifically, we show improved empirical results on CIFAR10, CIFAR100 and ImageNet using state-of-the-art {\it convolutional neural networks} (CNNs) and BERT\textsubscript{BASE} architecture for the MNLI, QQP, QNLI, MRPC, SST-2 and RTE benchmarks. 

\ignore{
We present a selective sampling method designed to accelerate the training of deep neural networks. To this end,  we introduce a novel measurement,  the {\it minimal margin score} (MMS), which measures the minimal amount of displacement an input should undergo until its predicted classification is switched. For multi-class linear classification,  the MMS measure is a natural generalization of the margin-based selection criterion, which has been thoroughly studied in the binary classification setting.  In addition, the MMS measure provides an interesting insight into the progress of the training process and can be useful for designing and monitoring new training regimes. We demonstrate empirically that when training commonly used deep neural network architectures for popular image classification tasks, there is a marked and substantial acceleration in the training process.  The efficiency of our method is compared against standard training procedures, and against commonly used selective sampling alternatives: Hard negative mining selection, and Entropy-based selection.
Finally, we demonstrate an additional speedup when we adopt a more aggressive learning-drop regime while using the MMS selective sampling method.
}

\end{abstract}

\section{Introduction}
\label{sec:intro}

Over the last decade, deep neural networks (DNNs) have become the {\it machine learning} method of choice in a variety of applications,  demonstrating outstanding performance, often close to or above the human-level.
Despite their success, some researchers have shown that neural networks can generalize poorly even with small data transformations \cite{azulay2018deep} as well as overfit to arbitrarily corrupted data \cite{zhang2017multi}. Additionally, problems such as adversarial examples \cite{szegedy2013intriguing, goodfellow2014explaining}, which cause neural networks to misclassify slightly perturbed input data, 
%can be a big issue towards
can be a source of concern in real-world deployment of models. 
These challenges raise the question as to whether properties that enabled classical machine learning algorithms to overcome these problems can be useful in helping DNNs resolve similar problems. 
%Specifically for linear classifiers case of {\it Support Vector Machine} (SVM)
%\cite{cortes1995support}, it has been evident that a classifier with large margin over different classes in the training data produced better generalization results as well as robustness for input perturbations \cite{bousquet2001algorithmic} on unseen data. 
Specifically, \cite{schapire1998boosting} introduced margin theory to explain boosting resistance to over-fitting. 
%Later, Reyzin & Schapire (2006) conjectured that the margin distribution, rather than the minimum margin, plays a key role in being empirically resistant to overfitting problem; this has been finally proved by Gao & Zhou (2013).
\ignore{
Furthermore, will the large margin principle, i.e., maximizing the smallest distance from the instances to the classification boundary in the feature space, that has played an important role in theoretical analysis of generalization, and helped to achieve remarkable practical results \cite{cortes1995support}, attain the same results for DNNs? 
}
Furthermore, the large margin principle, i.e., maximizing the smallest distance from the instances to the classification boundary in the feature space, has played an important role in theoretical analysis of generalization, and helped to achieve remarkable practical results \cite{cortes1995support},  
%It has been evident that a classifier with a large margin over different classes in the training data produces better generalization results 
as well as robustness for input perturbations \cite{bousquet2001algorithmic} on unseen data. 
\ignore{
Of particular interest to our study are the extensions to multi-class classification: multi-class perceptron (see Kesler’s construction,~\cite{duda1973pattern}); multi-class SVM \cite{vapnik1998statistical}; multi-class margin distribution \cite{zhang2017multi}, and the mistake-bound for multi-class linear sperability that scales with $(R/\gamma)^2$, where $R$ is the maximal norm of the samples in the feature space, and the $\gamma$ is the margin~\cite{crammer2003ultraconservative}.
}
Can the application of the large margin principle in DNNs lead to similar results? 

Although computation of the actual margin in the input space of DNNs is intractable, studies show that the widely used cross-entropy loss function is by itself a proxy for converging to the maximal margin \cite{soudry2018implicit}. To date, this was only demonstrated for linear models that, similarly to SVM, have a theoretical guarantee for maximal margin convergence \cite{rosset2004margin}. No such assurance for non-linear DNNs, their being being highly non-convex, has been offered.

Recently, \cite{jiang2019predicting} developed a measure for predicting the generalization gap~\footnote{The difference in accuracy between training and testing performance.} in DNNs that leverages {\it margin distribution}~\footnote{The distribution of distances to the {\it decision boundaries}.} \cite{garg2002generalization} as a more robust assessment for the margin notion in DNNs. In their work, they also point out that this measure can be used as an auxiliary loss function to achieve better generalization.

\ignore{
Adopting the idea of large margin principle from the SVM, \cite{jiang2019predicting} developed a measure for predicting the generalization gap in DNN (difference in accuracy between the raining set and testing set) which to a large extent leverages {\it margin distribution}~\footnote{The distribution of distances to the {\it decision boundaries}.} \cite{garg2002generalization} as a more robust assessment for the margin notion in deep neural networks. In their work, they also point out that this measure can be used as an auxiliary loss function to achieve better generalization.
}
%Recently, based on [2], [15], the authors of [19] argues that the margin distribution rather than a single margin is really crucial for improving generalization ability.

In the present study, we extend the aforementioned ideas and present a novel regularization term, which we denote as {\it Multi-Margin Regularization} (MMR), which can be added to any existing loss function in DNNs.
We derive the regularization term starting from the binary case of large margin classification and generalize it to the multi-class case.
%and then to the maximal margin principle in DNNs. 
%In contrast to other conventional methods, our method relies on the margin induced by classifiers attained from the true class and the second maximal class score. 
This regularization term aims at increasing the margin induced by classifiers attained from  the true class and its most competitive class. By summing over the margin distribution we compensate for class imbalance in the regularization term.
%Furthermore, since we apply our method to the feature space and since it is dynamic in the course of training of neural networks, we address this by scaling our formulation by the radius of the feature space, i.e., the maximal norm of the samples in the feature space. 
Furthermore, due to the dynamic nature of feature space representation when training neural networks, we scale our formulation %by the ever changing radius, $\| \phi_{max}\|$, in the feature space, i.e. the maximal norm of the samples in the feature space. 
by the ever-changing maximal norm of the samples in the feature space, $\| \phi_{max}\|$. 

We empirically show that applying this regulizer on the output layer of various DNNs, in different classification tasks and from different domains, is sufficient to obtain a substantial improvement in accuracy. In particular, we achieve valuable accuracy improvement in numerous image and text classification tasks, including CIFAR10, CIFAR100, ImageNet, MNLI, QQP and more.

In fact, our contribution is twofold. Alongside improving generalization performance using a new regularization scheme, we leverage the large margin principle to improve convergence during training using a selective-sampling scheme and assisted by a measure that we call the {\it Minimal Margin Score} (MMS). 
Essentially, MMS measures the distance to the decision boundary of the two most competitive predicted labels. 
%Specifically, we suggest a method 
This measure, in turn, is used 
to select, at the back-propagation pass, only those instances that accelerate %the training
convergence, thus speeding up the entire training process. 
It is worth noting that
our selection criterion is based on computations that are 
%calculated in any event as  
an integral part of the forward pass, thus taking advantage of the "cheaper" inference computations. Lastly, we empirically show that using the MMS selection scheme with a faster learning-rate regime can improve, to a large extent, the convergence process.

\subsection{Previous Approaches}
\ignore{
Large margin principle has proven to be fundamentally important in the history of {\it machine learning}
%linear classifiers 
\cite{rosset2004margin}. It has also been shown to correlate with better generalization properties. \cite{schapire1998boosting} demonstrated these ideas on bagging and boosting techniques. A true margin, however, is intractable in DNNs.
}

The large margin principle has proven to be fundamentally important in the history of machine learning. While most of the efforts revolved around binary classification, extensions to multi-class classification were also suggested, e.g., multi-class perceptron (see Kesler’s construction,~\cite{duda1973pattern}), multi-class SVM \cite{vapnik1998statistical} and multi-class margin distribution \cite{zhang2017multi}.
Margin analysis have also been shown to correlate with better generalization properties~\cite{schapire1998boosting}. Of particular interest to our study is
the mistake-bound for multi-class linear separability that scales with $(R/\gamma)^2$, where $R$ is the maximal norm of the samples in the feature space, and $\gamma$ is the margin~\cite{crammer2003ultraconservative}.

Computing the actual margin in DNNs, though, is intractable. \cite{soudry2018implicit} proved that cross-entropy loss in linear DNNs, together with {\it stochastic gradient descent} (SGD) optimization, converges to a maximal margin solution, but it cannot ensure a maximal margin solution in nonlinear DNNs. \cite{Sun2015LargeMD} affirmed that cross-entropy alone is not enough to achieve the maximal margin in DNNs and that an additional regularization term is needed.  

Several works addressed the large margin principle in DNNs. \cite{elsayed2018large} presented a multi-class linear approximation of the margin as an alternative loss function. They applied their margin-based loss at each and every layer of the neural network. Moreover, their method required a second order derivative computation due to the presence of first order gradients in the loss function itself. Explicit computation of the second order gradients for each layer of the neural network, however, can be quite expensive, especially when DNNs are getting wider and deeper. To address this limitation, they used a first order linear approximation to deploy their loss function more effectively. 
Later, \cite{jiang2019predicting} presented a 
%measure, based on the SVM margin 
margin-based measure
that strongly correlates with the generalization gap in DNNs. Essentially, they measured the difference between the training and the test %capabilities of a neural network by obtaining statistical information based on 
performances of a neural network using statistics of
the marginal distribution \cite{garg2002generalization}. \cite{sokolic2017robust} used the input layer to approximate the margin via the Jacobian matrix of the network and showed that maximizing their approximations leads to a better generalization. In contrast, we show that applying our margin-based regularization to the output layer alone achieves substantial improvements. 

In addition to better generalization, we show that the large margin principle can also be used to accelerate the training of DNNs. Accelerating the training process is a long-standing challenge that has already been addressed by quite a few authors \cite{bengio2008adaptive, salimans2016weight, goyal2017accurate}. Specifically, we seek to highlight faster convergence via selective sampling.
%a sample selection approach. 
To date, the most notable sample selection approach is probably hard negative mining \cite{schroff2015facenet}, where samples are selected by their loss values. The underlying assumption is that samples with higher losses have a significant impact on the model. Recent works employ selection schemes that examine the importance of the samples ~\cite{alain2015variance,loshchilov2015online}. During training, the samples are selected based on their gradient norm, which in turn leads to a variance reduction in the stochastic gradients;
see also~\cite{katharopoulos2018not}.
%\cite{katharopoulos2018not} used selective sampling to choose the training samples that reduce the gradient variance. 
Our selection method, though, utilizes uncertainty sampling, where the selection criterion is the proximity to the decision boundary, and we use the MMS measure to score the examples.

\section{Margin Analysis for Binary and multi-class Classification}
\label{sec:MMS}

\ignore{
The large margin principle is traditionally presented in the context of a shallow linear classifier \cite{XX}. The large margin is the core principle behind the SVM classifier. It was shown that a maximizing the margin between the data samples and the decision boundary maximizes the generalization capabilities of the classifier \cite{XX}. In the original work of Vapnik \cite{cortes1995support} the large margin principle was applied to the closest data points  to the boundary (support vectors), while later works \cite{zhang2019optimal} extends the large margin principle to the mean and variance of the distances. 

We start our discussion with the preliminary derivation of the classical SVM and its maximal margin principle. Next we show its generalization to multi-class and the necessary ingredient we must comprise in order to adapt it to the deep and multi-class scenarios.  
}
Consider a classification problem with two classes ${\cal Y} \in \{ +1,-1 \}$. 
We denote by ${\cal X} \in {\cal R}^d$ the input space. 
Let $f(\w^T \x + b)$ be a linear classifier, where $\x \in {\cal X}$ and 
$$
f(z)= \left\{
\begin{array}{ll}
 +1 &~~\mbox{if}~~ z \geq 0 \\
 -1	&~~\mbox{otherwise} 
 \end{array}
 \right.
$$

The classifier is trained using a set of examples $\{(\x_1,y_1),(\x_2,y_2),\cdots, (\x_n,y_n)\} \in ({\cal X} \times {\cal Y})^m$ 
where each example is sampled identically and independently from an unknown distribution ${\cal D}$ over ${\cal X} \times {\cal Y}$. 
The goal is to classify correctly new samples drawn from ${\cal D}$.
%, i.e. $y f(\w^T \x) \geq 0,~~(\x,y)\in{\cal D}$.
% In order to train the classifier, we require that each example from the training set will be classified correctly, namely: $y_i f(\w^T \x_i) \geq 0$. However, its capability to correctly classify new data, drawn from ${\cal D}$, is shown to be correlated with minimal margin of the training set to the decision boundary.

Denote by $\ell$ the (linear) decision boundary of the classifier
\begin{equation}
\label{eq:ell}
\ell = \{ \x ~|~ \w^T \x + b = 0\} 
\end{equation}
The geometric distance of a point $\x$ from $\ell$ is given by %\cite{Cortes&Vapnik}
\begin{equation}
\label{eq:d_i}
d(\x) = \frac{\w^T \x +b}{\| \w \|}
\end{equation}
For a linearly separable training set,
%we can find many consistent classifiers, 
there exist numerous consistent classifiers, 
i.e., classifiers that classify all examples correctly. Better generalization, however, is achieved by selecting the classifier that maximizes the margin $\hat d$,
%However, in order to increase the capacity (generalization capability) of $f$, 
%the SVM with hard margin aims at maximizing the minimum margin $\hat d$
$$
\arg \max_{\w,b} \hat d ~~~\mbox{s.t.}~~~y_i \frac{\w^T \x_i +b}{\| \w \|} \geq \hat d,  ~~~~\forall i=1,\cdots,m
$$
This optimization is redundant with the length of $\w$ and $b$. 
%Namely if $\w^*,b^*$ is the optimal solution, then so  $\alpha \w^*, \alpha b^*$.
%In order to remove the redundancy,  we impose $y_i (\w^T \x_i +b) \geq 1$ which is translated into the following equivalent minimization problem~\cite{cortes1995support}
Imposing $y_i (\w^T \x_i +b) \geq 1$ removes this redundancy and results in the following equivalent minimization problem~\cite{cortes1995support}:
$$
\min_{\w,b} \|\w\|^2 ~~~\mbox{s.t.}~~~ y_i (\w^T \x_i +b) \geq 1,
~~~~\forall i=1,\cdots,m $$
%The above optimization enforces all samples to be classified correctly (with margin no less than 1). 
%For noisy data, or a data set that is not linearly separable, the set of linear constraints are relaxed and substituted with the hinge loss, i.e. soft margin 
To handle noisy and linearly inseparable data, the set of linear constraints can be relaxed and substituted by the hinge loss,
\begin{equation}
\label{eq:min}
    \min_{\w,b} \|\w\|^2+ \lambda \sum_i \max (0, 1- y_i (\w^T \x_i +b))
\end{equation}
The left term in Formula \ref{eq:min} is the {\em regularization} component and it promotes increasing of the margin between the data points and the decision boundary.  The right term of the formula is the {\it empirical risk} component, imposing correct classifications on the training samples. 
The two terms employ two complementary forces; the former improves the generalization capability while the latter ensures the classification will be carried out correctly.

\begin{figure}[bth]
% \vspace{-1em}
    \centering
    \includegraphics[width=4.5cm]{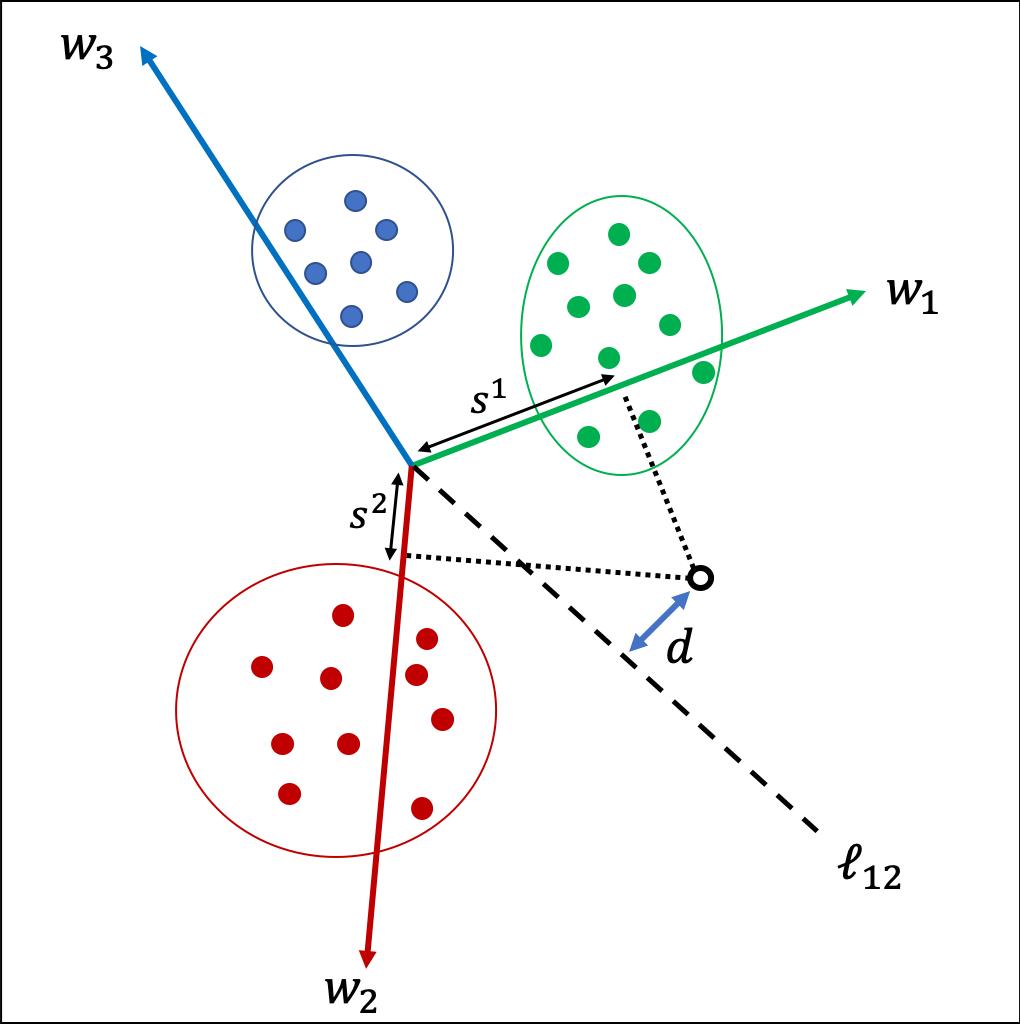}
    \caption{Illustrative example of a bi-class decision boundary. }
    \label{fig:mms_new}
\end{figure}

Next, we extend the large margin principle to the multi-class case. Let us assume we have a classification problem with $n$ classes, ${\cal Y} \in \{1,\cdots,n \}$, and a set of $m$ training samples: $\{ (\x_i, y_i)\} \in ({\cal X} \times {\cal Y})^m$. 
We now assign a score to each class: $s_i: {\cal X}\rightarrow \mathbb{R}, ~~\forall~i=1..n$. For a linear classification, the $j^{th}$ score of point $i$ is:
$$
s_j(\x_i) = \w_j^T \x_i + b_j
$$
The predicted class is chosen by the maximal score attained over all classes,
$$
{\hat y}_i = \arg \max_j s_j(\x_i)
$$
For any two classes, $(p,q) \in {\cal Y} \times {\cal Y}$, the decision boundary between these classes is given by (see Figure~\ref{fig:mms_new}):
$$
\ell_{p,q} = \{\x ~|~ s_p(\x)=s_q(\x) \} =
\{ \x ~|~  \w_p^T \x + b_p = \w_q^T \x + b_q \}.
$$
Denoting $\w_{p,q}=\w_p -\w_q$ and $b_{p,q}=b_p -b_q$, the decision boundary $\ell_{p,q}$ can be rewritten as:
$$
\ell_{p,q} = \{\x ~|~ \w_{p,q} \x + b_{p,q} =0 \} 
$$
which is similar to the binary case in Equation~\ref{eq:ell} where $\w_{p,q}$ replaces $\w$ and $b_{p,q}$ replaces $b$. 
Similarly to Equation~\ref{eq:d_i}, the geometric distance of a point $\x$ from $\ell_{p,q}$ is
\begin{equation}
\label{eq:dpq}
d_{p,q}(\x) =  \frac{\w_{p,q}^T \x +b_{p,q}}{\| \w_{p,q} \|}
\end{equation}

For point $\x_i$, denote by  $s_{y_i}(\x_i)$ the score for the true class and by $s_{m_i}(\x_i)$ the maximal score attained for the non-true classes, 
i.e., $m_i = \arg \max_{j \neq y_i} s_j(\x_i)$.
%By the above equation, we can define the distance of $\x_i$ from $\ell_{y_i,m_i}$:
Class $m_i$ is the {\it competitive} class vis-\`{a}-vis $y_i$. The boundary decision between $y_i$ and its competitive class is $\ell_{y_i,m_i}$ whose geometric distance to $\x_i$ is
\begin{equation}
\label{eq:d2}
d_{y_i,m_i}(\x_i) =  \frac{\w_{y_i,m_i}^T \x_i +b_{y_i,m_i}}{\| \w_{y_i,m_i} \|} 
\end{equation}
%where $\w_{y_i,m_i}=\w_{y_i}-\w_{m_i}$ and $b_{y_i,m_i}=b_{y_i}-b_{m_i}$.
Note that $d_{y_i,m_i}(\x_i)$ is non-negative if the classification is correct ($s_{y_i}(\x_i) \geq s_{m_i}(\x_i)$ ) and negative otherwise.

%We continue the derivations similarly to the binary case. For the multi-class case, Equation~\ref{eq:min} can be generalized into the following optimization problem:
For the multi-class case, Equation~\ref{eq:min} can be generalized to the following optimization problem,
\begin{equation}
\label{eq:optim2}
    \min_{W,\b}  \sum_i \|\w_{y_i,m_i}\|^2+ \lambda \sum_i \max (0, 1- (\w_{y_i,m_i}^T \x_i +b_{y_i,m_i}))
\end{equation}
where the optimization is over $W \doteq  \{\w_1,\cdots,\w_n\}$, and $\b \doteq \{b_1,\cdots,b_n\}$. Here too, the left-hand term is the regularization penalty while the right-hand term represents the empirical risk with a hinge loss. The regularization term aims to increase the margin between the true class and its competitive class. Note, though, that the summation is over the margin distribution ($i$ is the instance index). If the instances are evenly distributed over the classes, then this is equivalent to summation over the classes. Otherwise, this summation compensates for class imbalance in the regularization term.
\ignore{
It is worth mentioning that in contrast to the common $L_2$ regularization where the Frobenius norm of $W$ is minimized: $\sum_{i=j}^n \|\w_j\|^2$, the above scheme suggests to minimize $\sum_i \|\w_{y_i,m_i}\|^2$ which is a different regularization objective. For a pair of classes $(i,j)$, the standard $L_2$ regularization minimizes $\|\w_i\|^2+\|\w_j\|^2$ while in the suggested scheme we minimize $\|\w_i\|^2+\|\w_j\|^2 - \w_i^T \w_j$.
}

\section{Large Margin in DNNs}
\label{MMP_details}
Applying the above scheme directly to DNNs poses several problems. First, these networks employ a non-linear mapping from the input space into a representation space: 
%$\phi_i= m(\x_i): {\cal X} \rightarrow {\Phi}$.  
$\phi_i= F(\x_i, \theta): {\cal X} \rightarrow {\Phi}$, where $\theta$ are the network parameters.  
The  vector $\phi_i$  can be interpreted as a feature vector based on which the last layer in a DNN calculates the scores for each class via a fully-connected layer,  $s_j(\phi_i)=\w_j^T \phi_i + b_j$. 
Maximizing the margin in the input space ${\cal X}$, as suggested in \cite{sokolic2017robust}, requires back-propagating derivatives downstream the network up to the input layer, and calculating distances to the boundary up to the first order of approximation. In highly non-linear mappings, this approximation loses accuracy very fast as we move away from the decision boundary. 
Therefore, we apply the large margin principle in the last layer, where the distances to the decision boundary are Euclidean in the feature space $\Phi$:
\begin{equation}
\label{eq:d3}
    d_{y_i,m_i}(\phi_i) =  \frac{\w_{y_i,m_i}^T \phi_i +b_{y_i,m_i}}{\| \w_{y_i,m_i} \|} 
\end{equation}

The second problem stems from the fact that in Equation~\ref{eq:d2} the input space ${\cal X}$ is fixed along the course of training while the feature space $\Phi$ in Equation~\ref{eq:d3} is constantly changing. Accordingly, maximizing the margins in Equation \ref{eq:d3} can be trivially attained by scaling up the space $\Phi$.  Therefore, the feature space $\Phi$ must be constrained. In our scheme, we divide Equation \ref{eq:d3} by $\| \phi_{max}\|$, 
the maximal norm of the samples in the feature space, of the current batch.
%which is the sample with the maximal feature magnitude at the current batch. 
This ensures that scaling up the feature space will not increase the distance in a free manner. The proposed formulation is translated, similarly to Equation~\ref{eq:optim2}, into the following optimization problem 
\begin{equation}
\label{eq:d4}
\min_{W,\b}  \sum_i {\cal R}_i + \lambda \sum_i {\cal C}_i
\end{equation}
where 
$${\cal R}_i = \|\w_{y_i,m_i}\|^2 \|\phi_{max}\|^2 $$
denotes the margin regularization term, and ${\cal C}_i$ is the empirical risk term.
% \begin{equation}
% \label{eq:hinge}
% {\cal C}_i = \max (0, 1- (\w_{y_i,m_i}^T \phi_i + b_{y_i,m_i}))
% \end{equation}
% is the empirical risk term composed of hinge loss.
% \begin{equation}
% \label{eq:optim3}
%     \min_{W,\b}  \sum_i \|\w_{y_i,m_i}\|^2 \|\phi_i\|^2 + \lambda \sum_i \max (0, 1- (\w_{y_i,m_i}^T \x_i +b_{y_i,m_i}))
% \end{equation}
\ignore{ 
 The last issue in implementing the above formulation in DNN is that hinge loss provides inferior results compared to  the cross-entropy. This was demonstrated over a large set of problems \cite{XX} and in fact, the common practice today is to apply only cross-entropy criterion in DNN solving multi-class problems.  However, this issue can be easily mitigated by replacing the hinge loss in Equation~\ref{eq:hinge} with  cross entropy: 
\begin{equation}
{\cal C}_i = - \log (P_{y_i})
\end{equation}
 where $P_{y_i}$ is the conditional probability of the true label $y_i$ as obtained from the network after the softmax layer.  Similarly to  hinge loss, the cross entropy will promote the correct classification while the regularization term will maximize the margin. 
 }
 While for SVM, hinge loss is commonly used, in DNNs the common practice is to use cross-entropy
 %, i.e. ${\cal C}_i = - \log (P_{y_i})$, 
$$
{\cal C}_i = - \log (P_{y_i})
$$
 where $P_{y_i}$ is the probability of the true label $y_i$ obtained from the network after the softmax layer: 
 $$
 P_{y_i}=\frac{e^{s_{y_i}(\x_i)}}{\sum_j e^{s_j(\x_i)} }
 $$
 Similarly to hinge loss, cross-entropy will strive for correct classification while the regularization term will maximize the margin. For the rest of this paper we denote ${\cal R}_i$ as the {\it multi-margin megularization} (MMR).
 
 Note that the regularization term in this scheme is different from the weight decay commonly applied in deep networks. First, here, the minimization is applied over the $\w$ differences of:  $\|\w_{y_i,m_i}\|^2=\|\w_{y_i} -\w_{m_i}\|^2$. Additionally, the regularization term is multiplied by the $\|\phi_{max}\|$.
 Lastly, the regularization term is implemented only at the last layer. 
 %and it does not applied to each layer independently.  

\section{Accelerating Training Using Minimal Margin Score Selection}
\label{sec:SelectSample}

\iffalse {The traditional active learning schemes are based on querying labels from a limited budget in an informative and adaptive manner. In a wide range of cases, these schemes offer an exponential reduction in label complexity and therefore an exponential reduction in computation. In more complex and non-convex models such as CNNs, we cannot make any assumptions with respect to the entire model hypothesis space, thus we will direct our focus on the part were the data become more and more linearly separable due to the optimization process, namely, the last fully-connected layer which is also the final classification layer.

We further leverage these concepts and apply our selection based on the samples that are close to the margin in the multi-class scenario. 
}
\fi
\iffalse {
As mentioned above, our method is based on the evaluation of the minimal amount of displacement a training sample should undergo until its predicted classification is switched. We call this measure {\em minimal margin score} (MMS). This measure depends on the best and the 2nd best scores achieved per sample. 
Our measure was inspired by the margin-based quantity suggested by ~\cite{jiang2019predicting} for predicting the generalization gap of a given network. However, in our scheme, we apply our measure to the output layer, and we calculate it linearly with respect to the input of the last layer. Additionally, unlike ~\cite{jiang2019predicting}, we do not take into consideration the true label, and our measure is calculated based on the best and the 2nd best NN scores. }
\fi

We continue to leverage the principle of large margin in neural networks to address the computational limitations in real-world applications, specifically in the selective sampling scheme. We show that by selecting samples that are closer to the margin in the multi-class setting during the forward pass, we achieve better convergence and speed-up during the training process.
To this end, we evaluate our suggested method on CIFAR10 and CIFAR100 using ResNet-44 \cite{he2016deep} and WRN-28-10 \cite{zagoruyko2016wide} architectures. To demonstrate the effectiveness of our selection more vigorously, we apply a faster learning-rate (LR) regime than those suggested in the original papers. 

In principle, our selection method is based on the evaluation of the minimal amount of displacement a training sample should undergo until its predicted classification is switched. We call this measure the {\em minimal margin score} (MMS). This measure depends on the highest and the second highest scores achieved per sample. 
Similarly to our margin-based regularization, we apply our measure only to the output layer, and calculate it linearly with respect to the input of the last layer. Additionally, unlike ~\cite{jiang2019predicting}, we do not take into consideration the true label, i.e., our measure is calculated based solely on the highest and the second highest neural network scores.

As shown in Figure~\ref{fig:mms_new}, a multi-class classification problem is composed of three classes, Green, Red, and Blue, along with three linear projections, $\w_1, \w_2,$ and $\w_3$, respectively. The query point is marked by an empty black circle. The highest scores of the query point are $s^1$ and $s^2$ (assuming all biases are 0's), where $s^1>s^2$ and $s^3$ are negative (not marked). Since the two highest scores are for the Green and Red classes, the distance of the query point to the decision boundary between these two classes is $d$. The magnitude of $d$ is the MMS of this query point.

\ignore{
\begin{figure}[!bht]
% \vspace{-1em}
    \centering
    \includegraphics[width=5cm, height=5cm]{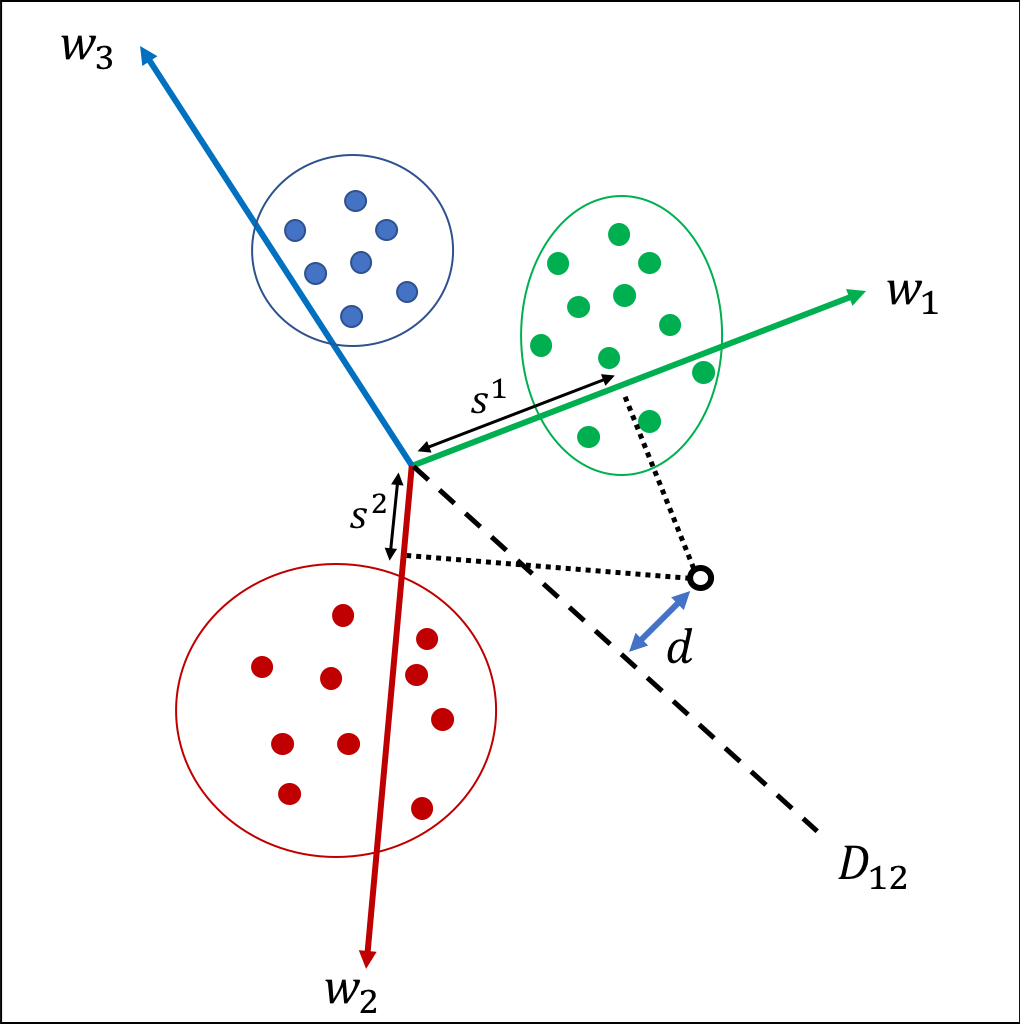}
    \caption{Illustrative example of the MMS measure. }
    \label{fig:mms}
\end{figure}
}

Formally, let $\mathcal{X} = \{\x_1,...,\x_{m}\}$ be a large set of samples and $\phi_i=F(\x_i; \theta) \in \Phi$ be the input to the last layer of the neural network. 
 Assume we have a classification problem with $n$ classes. At the last layer, the classifier $s$ consists of $n$ linear functions: $s_j: \mathcal{\phi} \rightarrow \mathbb{R}$ for $j=1\dots n$ where $s_j$ is a linear mapping $s_j(\phi)=\w_j^T \phi + b_j$.
Denote the sorted scores of $\{s_j(\phi_k)\}_{j=1}^n$ by ${\cal S}=(s_k^{j_1},s_k^{j_2},\cdots,s_k^{j_n})$, where $s_k^{j_p} \geq s_k^{j_{p+1}}$ and  $s_k^{j_p}=s_{j_p}(\phi_k)$.
The classifier $s_{j_1}(\phi_k)$ gives the highest score and $s_{j_2}(\phi_k)$, the second highest score. The {\em decision boundary} between classes $j_1$ and $j_2$ is defined as: 
$$\ell_{12}=\{\phi ~|~ s_{j_1}(\phi)=s_{j_2}(\phi) \}$$
Using this definition, the confidence of the predicted label $j_1$ of point $\phi_k$ is determined  by the distance $d_k$ of $\phi_k$ to the decision boundary $\ell_{12}$.
Following Equation~\ref{eq:d3}, it is easy to show  that
\begin{equation}
d_k = \frac{s_k^{j_1} -s_k^{j_2} }{\|\w_{j_1} - \w_{j_2}\|}
\end{equation}
The distance $d_k$ is the MMS of point $\x_k$. The larger $d_k$, the more confident we are about the predicted label. Conversely, the smaller $d_k$, the less confident we are about the predicted label $j_1$. Therefore, $d_k$ can serve as a confidence measure for the predicted labels. Accordingly, the best points to select for the back-propagation step are the points whose MMS are the smallest. Note that in contrast to the MMR, in this case we do not have to normalize the distance $d_k$ with $\| \phi_{max} \|$ because this normalization will not change the order of ${\cal S}$,
%and the point selection will remain unchanged. 
and thus the set of selected points will remain unchanged.

Our implementation consists of a generic, yet simple, online selective sampling method, 
%applied as a preliminary part of each training step. 
applied at the beginning of each training step.
Specifically, at each training step, we first apply a forward pass on a batch of points of size $B$, and obtain their respective scores.We then calculate their respective MMS measures, and select the $b$ samples ($b \ll B$) whose MMS measures are the smallest. The resulting batch of size $b$, in turn, is used for training the network. The selection process is repeated every training step, thus potentially selecting a new batch of points for training. The MMS-based training procedure is summarized in Algorithm \ref{alg:batch_select}.

% \begin{center}% Or use a figure environment and \centering
% \begin{minipage}{1.0\linewidth}
\begin{algorithm}
% \noindent\begin{minipage}{\textwidth}
\renewcommand\footnoterule{}
\begin{algorithmic}[1]
\REQUIRE 
Inputs $\mathcal{X}=\{\x_i\}_{i=1}^B$ ,\ $F(\cdot; \theta_0)$ \\%- Training model, b - batch size\\
\STATE{$t \gets 1$}
\REPEAT
\STATE{${\Phi} \gets F(\mathcal{X}; \theta_{t-1})$ \quad forward pass a batch of size B}
\STATE $MMS \gets d(\Phi)$ \qquad \qquad  \ \ \ \ \ \ \ calc. MMS 
\STATE ${ S} \gets \mbox{sort\_index}(MMS,b)$  \ \ \ \ \ \ store $b$ smallest scores
\STATE{$\mathcal {X}_b = \{\x_i | ~ i \in \mathcal{S}\}$ \qquad \qquad \ \ \ \ subset of $\mathcal{X}$ of size b}
\STATE{$\theta_t \gets sgd\_step(F( \mathcal{X}_b; \theta_{t-1}))$ \ \ back prop. batch of size b}
\STATE {$t \gets t+1$} 
\UNTIL{reaching final model accuracy}
\end{algorithmic}
\caption{MMS-based training}
\label{alg:batch_select}
% \end{minipage}
\end{algorithm}
% \end{minipage}
% \end{center}

\section{Experiments}
\label{sec:Exp}
In this section\footnote{All experiments were conducted using PyTorch; the code will be released on github upon acceptance of the paper.},
%is publicly available\ at \url{https://github.com/paper-submissions/mms-select}.}, 
%we report on our experiments that evaluated our MMR and the MMS selection methods for achieving a higher accuracy score as well as faster convergence, respectively. 
we report on the series of experiments we designed to evaluate the MMR's ability to achieve a higher accuracy score, and the MMS selection method's ability to achieve a faster convergence than the original training algorithms (the baseline) and data augmentation. %described by their authors. 
%For the selection scheme, we used uniform sampling from the dataset. We used several common datasets and neural network based models taken from vision and {\it natural language processing} (NLP) domains.
The experiments were conducted on commonly used datasets and neural network models, in the vision and {\it natural language processing} (NLP) realms.

\begin{table*}[ht]
\centering
\begin{tabular}{l l l l l l l}
\toprule{}\
      
Model & Dataset &  Baseline  & Our MMR & Change \\
\midrule
ResNet-44 \cite{he2016deep} & CIFAR10 & 93.22\% & 93.83\% & \bf{9.00\%}   \\

VGG \cite{simonyan2014very} & CIFAR10 & 93.19\% & 93.34\% & 2.20\%   \\

WRN-28-10 + auto-augment + cutout \cite{zagoruyko2016wide} & CIFAR100 & 82.51\% & 83.52\% & \bf{5.77\%}   \\

VGG + auto-augment + cutout & CIFAR100 & 73.93\% & 74.19\% & 1.00\%   \\

MobileNet \cite{howard2017mobilenets} & ImageNet & 71.17\% & 71.44\% & 0.94\%   \\

\midrule

  & QNLI & 91.06\% & 91.48\% & \bf{4.70\%}   \\

  & SST-2 & 92.08\% & 92.43\% & \bf{4.42\%}   \\

BERT\textsubscript{BASE} \cite{devlin2018bert}  & MRPC & 90.68\% & 91.43\% & \bf{8.05\%}   \\

  & RTE & 68.23\% & 69.67\% & \bf{4.53\%}   \\

  & QQP & 87.9\% & 88.04\% & 1.16\%   \\

  & MNLI & 84.5\% & 84.70\% & 1.29\%   \\

\bottomrule
\end{tabular}
\caption{Test accuracy results. Top1 for CIFAR10/100 datasets. Any relative change in error over the baseline is listed in percentage, and improvements higher than 4\% are marked in bold. F1 scores are reported for QQP and MRPC. For MNLI, we report the average of the matched (with $\alpha=1e-5$) and miss-matched (with $\alpha=1e-6$) for both the baseline and our MMR.}
\label{table:val_accuracy}
%\vspace{-2em}
\end{table*}

Our experimental workbench is composed of CIFAR10, CIFAR100 \cite{krizhevsky2009learning} and ImageNet \cite{imagenet_cvpr09} for image classification; Question NLI (QNLI) \cite{wang2018glue}, MultiNLI (MNLI) \cite{williams2017broad} and Recognizing Textual Entailment (RTE) \cite{bentivogli2009fifth} for natural language inference; MSR Paraphrase Corpus (MRPC) \cite{dolan2005automatically} and  Quora Question Pairs (QQP) \cite{chen2018quora} for sentence similarity; Stanford Sentiment Treebank-2 (SST-2) \cite{socher2013recursive} for text classification.

\subsection{Image Classification}

\paragraph{CIFAR10 and CIFAR100.} These are image classification datasets that consist of $32 \times 32$ color images from 10 or 100 classes, consisting of 50k training examples and 10k test examples. The last 5k images of the training set are used as a held-out validation set, as suggested in common practice. For our experiments, we used ResNet-44 \cite{he2016deep} and WRN-28-10 \cite{zagoruyko2016wide} architectures. We applied the original hyper parameters and training regime using a batch-size of 64. In addition, we used the original augmentation policy as described in \cite{he2016deep} for ResNet-44, while adding cutout \cite{devries2017improved} and auto-augment \cite{cubuk2018autoaugment} for WRN-28-10. Optimization was performed for 200 epochs (equivalent to $156K$ iterations) after which baseline accuracy was obtained with no apparent improvement. 

\paragraph{ImageNet.} For large-scale evaluation, we used the ImageNet dataset \cite{imagenet_cvpr09}, containing more than 1.2M images in 1k classes. In our experiments, we used MobileNet \cite{howard2017mobilenets} architecture and followed the training regime established by \cite{goyal2017accurate} in which an initial LR of 0.1 is decreased by a factor of 10 in epochs 30, 60, and 80, for a total of 90 epochs. We used a base batch size of $256$ over four devices and $L_2$ regularization over weights of convolutional layers as well as the standard data augmentation.

\subsubsection{Improving accuracy via MMR}\
\newline
\ignore{
To examine our MMR scheme as described in detail in Section \ref{MMP_details}, we added it to the objective function as an additional regularization term. We used a trade-off $\alpha$ factor between the cross-entropy loss and the additional regularization as follows~\footnote{Equivalently to Equation~\ref{eq:d4} we used $\alpha=\frac{1}{\lambda}$ to multiply the regularization term by a small number and keep the scaling factor of ${\cal C}_i$ to be 1, to avoid gradient enlargement}:
}
The MMR was added to the objective function as an additional regularization term, where $\alpha$ is a trade-off factor between the cross-entropy loss and the regularization~\footnote{This formulation is equivalent to Equation~\ref{eq:d4}, where  $\alpha=\frac{1}{\lambda}$. It is preferred because it leads to multiplying the regularization term by a small number and keeping the scaling factor of ${\cal C}_i$ to be 1, thus avoiding gradient enlargement.}:
$$
\mathcal{L}(\theta) =  \alpha \sum_i {\cal R}_i + \sum_i {\cal C}_i
$$
%\newline
To find the optimal $\alpha$, we used a grid search and found that a linear scaling of $\alpha$ in the range of $[1e-5..1e-3]$ works best for  CIFAR10/100 and static $\alpha=1e-5$ works best for ImageNet. 
%\
%\newline
%\newline

\begin{figure}[t]
% \vspace{-1em}
    \centering
    \subfloat[\label{fig:lb}]{
     \includegraphics[width=0.8\columnwidth]{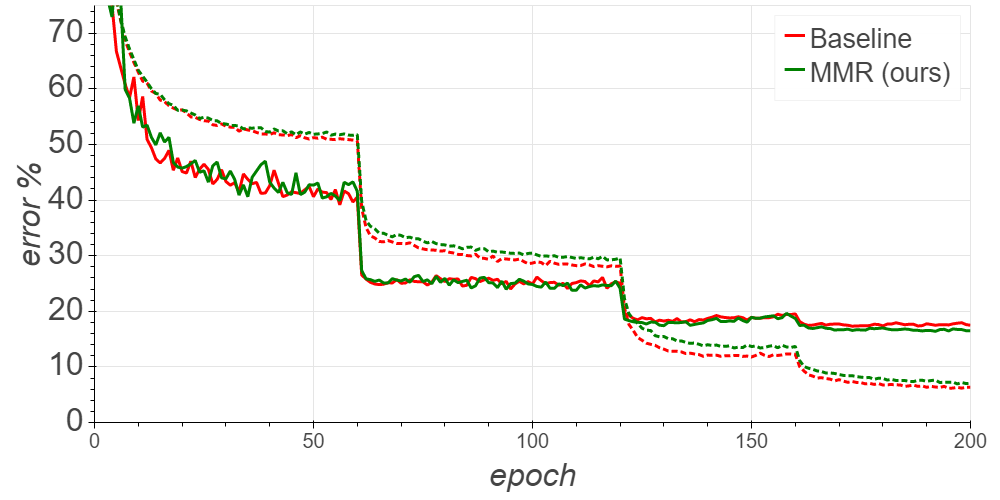} } \hfill
    % \qqqqqquad %add desired spacing between images, e. g. ~, \quad, \qquad, \hfill etc. 
      %(or a blank line to force the subfigure onto a new line)
    \subfloat[\label{fig:rb}]{
        \includegraphics[width=0.8\columnwidth]{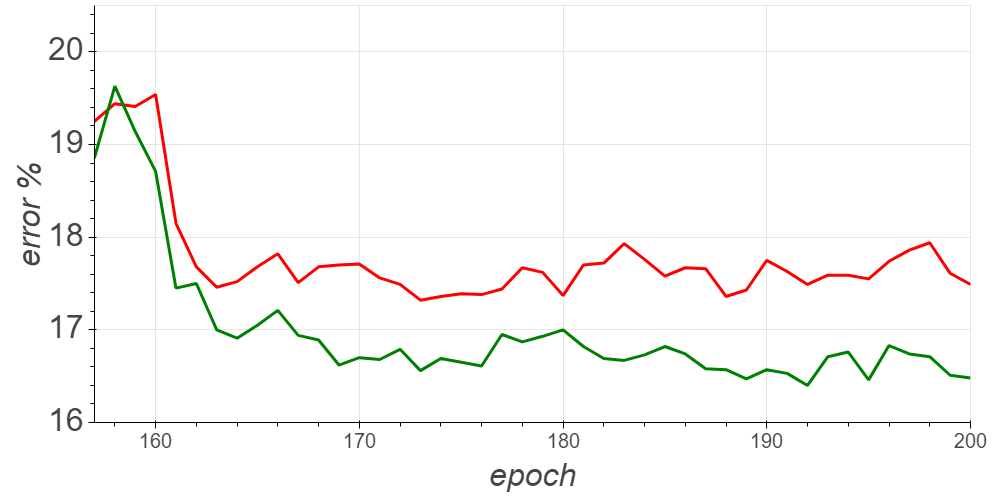} }
        % \caption{CIFAR100 on WRN28-10 val error with early drop regime}
     \caption{Training (dashed) and validation errors of CIFAR100 using the WRN28-10 neural network and comparing baseline training and our MMR approach. We use linear scale $\alpha$, starting with $1e-5$ up to $1e-3$.}
    \label{compare_cifar100_mmp_ref}
    % \vspace{-1em}
\end{figure}

Table~\ref{table:val_accuracy} demonstrates our final results when increasing the final model's accuracy on CIFAR10 and CIFAR100. Specifically, we managed to improve baseline accuracy in ResNet-44 from $93.22\%$ to $93.83\%$ and from $93.19\%$ to $93.34\%$ in VGG. We also see a relative change in error of $9.00\%$ and $2.20\%$, respectively, on the CIFAR10 dataset. Furthermore, we show a substantial decrease of $5.77\%$ in the error for CIFAR100 using the WRN-28-20 model (see Figure~\ref{compare_cifar100_mmp_ref}), raising its absolute accuracy by more than $1\%$.
Altogether, we observed a $2.67\%$ average decrease in error on all datasets.

\subsubsection{Convergence speedup via MMS selective sampling}
\begin{figure}[bht]
% \vspace{-1em}
    \centering
    % \qqquad
    % \begin{subfigure}
    %     \includegraphics[width=0.9\columnwidth]{Figures/resnet44_cifar10_train_select_mms_early_drop.png}
    %     % \caption{CIFAR10 on ResNet44 val error with early drop regime}
    %     \label{fig:1c}
    % \end{subfigure}
    % \begin{subfigure}
    %     \includegraphics[width=0.9\columnwidth]{Figures/resnet28_wide_cifar100_train_select_mms_early_drop.png}
    %     % \caption{CIFAR10 on ResNet44 val error with early drop regime}
    %     \label{fig:1d}
    % \end{subfigure}
    % \qqquad
    \subfloat[\label{fig:1e}]{
        \includegraphics[width=0.9\columnwidth]{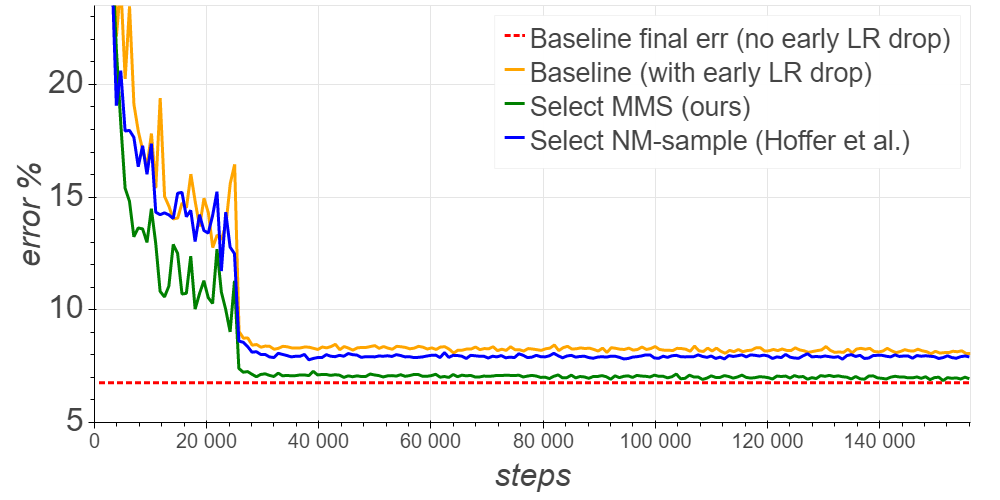} }
        % \caption{CIFAR10 on ResNet44 val error with early drop regime}
        
     %add desired spacing between images, e. g. ~, \quad, \qquad, \hfill etc. 
      %(or a blank line to force the subfigure onto a new line)
        \subfloat[\label{fig:1f}]{
        \includegraphics[width=0.9\columnwidth]{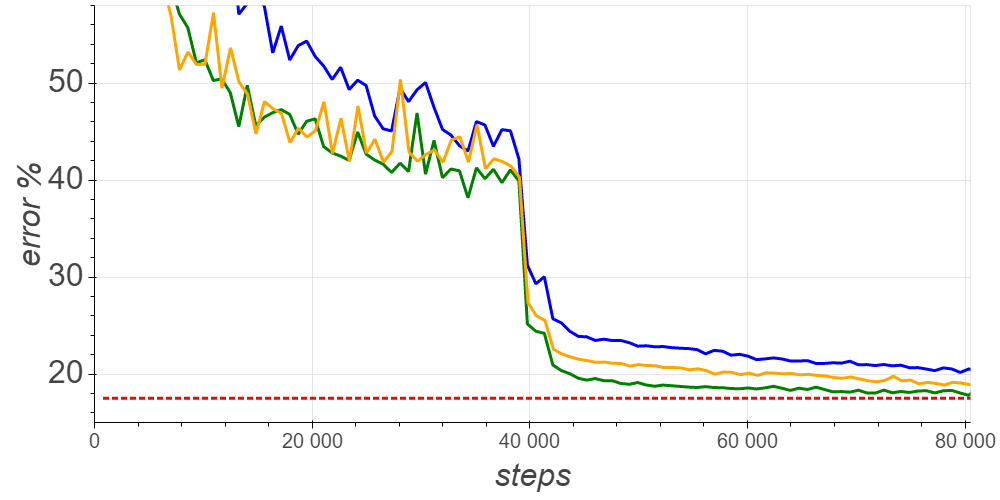} }
        % \caption{CIFAR100 on WRN28-10 val error with early drop regime}
     \caption{Validation errors of ResNet44, CIFAR10 (top) and WRN-28-10, CIFAR100 (bottom). We compared the baseline training, NM-sample selection (hard negative mining), and MMS (our) selection method using a faster regime. We ploted the regular regime baseline's final errors as a dotted line for perspective. \textbf{The MMS selection method achieves on par final test accuracy using fewer numbers of training steps}.}
    \label{compare_cifar10/100_early_drop}
    % \vspace{-1em}
\end{figure}

\
\newline
To test our hypothesis that using the MMS selection method we could accelerate training while preserving final model accuracy, we designed a new, more aggressive leaning-rate drop regime than the one used by the authors of the original paper. Figure~\ref{compare_cifar10/100_early_drop} presents empirical evidence supporting out hypothesis. 
%We compared our MMS method against hard-negative mining that prefers samples with low prediction scores as well as random selection~\footnote{Referred as baseline with and without an early LR drop.}.
We compared the results of our MMS method against random selection~\footnote{Referred to as baseline with and without an early LR drop.}, and against  
{\it hard-negative mining}
that prefers samples with low prediction scores~\cite{yu2018loss,schroff2015facenet}. For the latter, we used the implementation suggested by~\cite{hofferinfer2train}, termed "NM-sample", where the cross-entropy loss is used for the selection.
\ignore{
\paragraph{Hard negative samples.}  For this experiment, we implemented the "NM-sample" procedure \cite{hofferinfer2train}, similar to the classical methods for "hard-negative-mining" used by machine-learning practitioners over the years \cite{yu2018loss,schroff2015facenet}. We used the cross-entropy loss as a proxy for the selection, in which the highest loss samples were selected. 
}

\
\newline
For CIFAR10 and ResNet-44, we used the original LRs $\eta=\{0.1, 0.01, 0.001, 0.0001\}$ while decreasing them at steps $\{24992, 27335, 29678\}$, equivalent to epochs $\{{32, 35, 38}\}$ with a batch of size 64. As depicted in Figure \ref{compare_cifar10/100_early_drop} (top), 
we can see that our selection method indeed yields
validation accuracy extremely close to the one reached by the baseline training scheme, with considerably fewer training steps. Specifically, we reached $93\%$ accuracy after merely $44K$ steps (a minor drop of $0.25\%$ compared to the baseline). We also applied the early drop regime to the baseline configuration as well as to the NM-samples. Both failed to reach the desired model accuracy while suffering from a degradation of $1.57\%$ and $1.22\%$, respectively. 

Similarly, we applied the early LR drop scheme for
CIFAR100 and WRN-28-10, using $\eta=\{0.1, 0.02, 0.004, 0.0008\}$ and decreasing steps  $\{39050, 41393, 43736\}$ equivalent to epochs $\{50, 53, 56\}$, with batch of size 64. 
As depicted in Figure \ref{compare_cifar10/100_early_drop} (bottom), MMS accuracy reached $82.2\%$ with a drop of $0.07\%$ compared to the baseline, while almost halving the number of steps ($80K$ vs. $156K$). On the other hand, the baseline and the NM-sample schemes failed to reach the desired accuracy after we applied a similar early drop regime. 
For the NM-sample approach, the degradation was the most significant, with a drop of $2.97\%$ compared to the final model accuracy, while the baseline drop was approximately $1\%$.

These results are in line with the main theme of selective sampling that strives to focus training on more informative points. Training loss, however, can be a poor proxy for this concept. For example, the NM-sample selection criterion favors high loss scores, which obviously increases the training error, while our MMS approach selects uncertain points, some of which might be correctly classified. Others might be mis-classified by a small margin, but they are all close to the decision boundary, and hence useful for training.

\subsection{Natural Language Classification Tasks}
To challenge our premise that we could achieve a higher accuracy score, we examined our MMR on an NLP-related model and datasets. In particular, we used the BERT\textsubscript{BASE} model \cite{devlin2018bert} with 12 transformer layers, a hidden dimensional size of 768 and 12 self-attention heads. Fine-tuning was performed using the Adam optmizer as in the pre-training, with a dropout probability of 0.1 on all layers. Additionally, we used a LR of $2e-5$ over three epochs in total for all the tasks. We used the original WordPiece embeddings \cite{wu2016google} with a 30k token vocabulary. For our method, similarly to the image classification task, we also used the $\alpha$ factor in the objective function, and found via a grid search, $\alpha=1e-5$ to be the optimal~\footnote{We applied $\alpha=1e-6$ only to evaluate our method's accuracy with the miss-matched MNLI.}. 

We performed experiments on a variety of supervised tasks, specifically by applying downstream task fine-tuning on natural language inference, semantic similarity, and text classification. All these tasks are available as part of the GLUE multi-task benchmark \cite{wang2018glue}.

\paragraph{Natural Language Inference}
The task of natural language inference (NLI) or recognizing textual entailment means that when a pair of sentences is given, the classifier decides whether or not they contradict each other. Although there has been a lot of progress, the task remains challenging due to the presence of a wide variety of phenomena such as lexical entailment, coreference, and lexical and syntactic ambiguity. We evaluate our scheme on three NLI datasets taken from different sources, including transcribed speech, popular fiction, and government reports (MNLI), Wikipedia articles (QNLI) and news articles (RTE).

As shown in Table~\ref{table:val_accuracy}, our scheme using the regularization term outperformed baseline results on all the three tasks. We achieved absolute improvement of up to $1.44\%$ on RTE and a relative change in error of $4.53\%$. On QNLI and MNLI we also achieved higher scores of $91.48\%$ (accuracy) and $84.70\%$ (F1), outperforming the baseline results by $0.42\%$ and $0.2\%$, respectively.

\paragraph{Semantic Similarity}
This task involves predicting whether two sentences are semantically equivalent by identifying similar concepts in both sentences. 
It can be challenging for a language model to recognize syntactic and morphological ambiguity as well as compare the same ideas using different expressions or the other way around. 
We evaluated our approach on QQP and MRPC downstream tasks, outperforming baseline results as can be seen in Table \ref{table:val_accuracy}. On MRPC in particular, we achieved a relative change of more than $8\%$. 

\paragraph{Text Classification}
Lastly, we evaluated our method on the Stanford Sentiment Treebank (SST-2), 
which is a binary single-sentence classification task consisting of sentences
extracted from movie reviews with human annotations of their sentiment. 
Our approach outperformed the baseline by a relative error change of $4.42\%$.

Overall, applying MMR boosted the accuracy in all the reported tasks. This indicates that our approach works well for different tasks from various domains.

\section{Discussion}
\label{sec:Discussion}
We studied a multi-class margin analysis for DNNs and use it to devise a novel regularization term, the {\it multi-margin regularization} (MMR). Similarly to previous formulations, the MMR aims at increasing the margin induced by the classifiers, and it is derived directly, for each sample, from 
the true class and its most competitive class. The main difference between the MMR and common regularization terms is that MMR is scaled by $\| \phi_{max}\|$, 
which is the maximal norm of the samples in the feature space.
%by the ever changing radius, $\phi_{max}$, in the feature space, i.e. the maximal norm of the samples in the feature space, 
This ensures a meaningful increase in the margin that is not induced by a simple scaling of the feature space.  Additionally,  weight differences are minimized rather than the commonly used determinant or other norms of $W$. Lastly, MMR in formulated and performed over the margin distribution to compensate for class imbalance in the regularization term. The MMR can be incorporated with any empirical risk loss and  it is not restrictive to hinge loss or cross-entropy losses. And indeed, using MMR, we demonstrate improved accuracy over a set of experiments in images and text. 

Additionally, the multi-class margin analysis enables us to propose a selective sampling method designed to accelerate the training of DNNs. Specifically, we utilized uncertainty sampling,  where the criterion for selection is the distance to the decision boundary. To this end, we introduced a novel measurement, the {\it minimal margin score} (MMS), which measures the minimal amount of displacement an input should undergo until its predicted classification is switched. For multi-class linear classification, the MMS measure is a natural generalization of the margin-based selection criterion.
%, which was thoroughly studied in the binary classification setting.
\ignore{
We demonstrate a substantial acceleration for training commonly used DNN architectures for popular image classification tasks. 
The efficiency of our method is compared against the standard training procedures, and against commonly used selective sampling methods: Hard negative mining selection, and Entropy-based selection.
%other, heavily used, selective sampling methods.
Furthermore, we demonstrate an additional speedup when we adopt a more aggressive learning-drop regime.

Tracking the MMS measure throughout the training process provides an interesting insight into the training process. Figure \ref{compare_mms} demonstrates that the MMS measure monotonically increasing, even when training and validation errors are flattened. Subsequently, it flattens when training converges to the final model. This suggests that improvement in training can be obtained as long as there is uncertainty in the class labels. Furthermore, tracking the MMS measure may turn out to be useful for designing and monitoring new training regimes.
}

\ignore{
We demonstrate the efficiency of our method in acceleration of training of commonly used deep neural networks architectures and in popular image classification tasks and compare it to common against the standard training procedures, as well as other, heavily used, selective sampling criteria. Moreover, we demonstrate an additional speedup when we adopt a more aggressive learning-drop regime.
}

Our selection criterion was inspired by the active learning method, but our goal, to accelerate training, is different. Active learning is mainly concerned with labeling cost. Hence, it is common to keep on training until convergence, before turning to select additional examples to label. When the goal is merely acceleration, labeling cost is not a concern, and one can adapt a more aggressive protocol and re-select a new batch of examples at each training step.
\ignore{
However, such an approach is less efficient when it comes to acceleration. In such a scenario, we can be more aggressive; since labeling cost is not a concern, we can re-select a new batch of examples at each training step.
}
\ignore{
An efficient implementation is crucial for gaining speedup. Our scheme provides many opportunities for further acceleration. For example, fine-tuning the batch size in order to
%fine-tuning the sample size used to select and fill up a new batch, to 
balance between the selection effort conducted at the end of the forward pass, and the compute effort required for the back-propagation pass. This provides an opportunity to design and use dedicated hardware for the selection. In the past few years, custom ASIC devices that accelerate the inference phase of neural networks have been developed~\cite{goya,hofferinfer2train,jouppi2017datacenter}. Furthermore, in \cite{jacob2018quantization}, it was shown that using quantization for low-precision computation induces little or no degradation in accuracy. Low precision compute together with the fast inference provided by ASICs, provide opportunities for additional acceleration in the forward pass of our selection scheme. 
}

The MMS measure does not use the labels. Thus, it can be used to select samples in an active learning setting as well. 
Similarly to \cite{jiang2019predicting} the MMS measure can be implemented at other layers in the deep architecture. This enables selection of examples that directly impact training at all levels. The additional computation associated with such a framework makes it less appealing for the purpose of acceleration. For active learning, however, it may introduce an additional gain.
%, since the selection criterion chooses examples which are more informative for various layers.   
The design of a novel active learning method is left for further study.

\newpage
\bibliography{aaai21.bib}
\end{document}